%% file: rl-for-ecos.tex
\documentclass[runningheads]{llncs}
\usepackage[T1]{fontenc}
\usepackage{graphicx}
\usepackage[hidelinks]{hyperref} %
\usepackage{color}

\usepackage{cite}
\begin{document}
\title{Reinforcement Learning for Economic Policy:\\A New Frontier?}
\author{Callum Rhys Tilbury}
\authorrunning{Callum Rhys Tilbury}
\institute{School of Informatics, University of Edinburgh\\\email{callum@tilbury.co.za}}
\maketitle              %
\begin{abstract}
        Agent-based computational economics is a field with a rich academic history, yet one which has struggled to enter mainstream policy design toolboxes---plagued by the challenges associated with representing a complex and dynamic reality. The field of Reinforcement Learning~(RL), too, has a rich history, and has recently been at the centre of several exponential developments. Modern RL implementations have been able to achieve unprecedented levels of sophistication, handling previously unthinkable degrees of complexity. This review surveys the historical barriers of classical agent-based techniques in economic modelling, and contemplates whether recent developments in RL can overcome any of them.
\end{abstract}

\section{Introduction}
Modelling is an integral part of an economist's strategy for making sense of the world, and computer simulations have long been a part of the discipline. For example, as early as 1960, researchers were attempting to simulate the United States' economy in recession~\cite{Duesenberry1960}. Of course, much has changed in the nature of these simulations since then---such experiments were done by hand, with a small number of governing equations. Notwithstanding, the research into simulation certainly continued as computing power increased during the latter half of the 20th century, and bold predictions were made about the future of the field. In the early 1990s, Holland and Miller presented a seminal paper~\cite{holland1991artificial} proposing the development of ``Artificial Agents'' in economic modelling. The authors firmly believed that the discipline of Artificial Intelligence (AI) could offer new possibilities for economic analysis---formulating the idea of ``complex adaptive system'' models. Interestingly, in this early formulation, the agents were being modelled in the same way as in Reinforcement Learning (RL): \emph{adaptive} agents that have a value function for their various actions, and which try to maximize the cumulative reward over time, using Markov processes as a mathematical foundation. Out of these low-level, individual agent actions, emergent behaviour could be realised, the authors argued.

Perhaps due to the discipline of RL being relatively nascent at the time, research into adaptive agents for economic modelling mostly waned, but the excitement around rule-based agents certainly prevailed---where the agent's nature is hard-coded \emph{a priori}. A new schema of Agent-Based Computational Economics (ACE) arose, along with a wealth of literature on the topic, as detailed by Tesfatsion and Judd~\cite{tesfatsion2006handbook}. Prominent economists such as J. Doyne Farmer and Duncan Foley boldly declared that ``the economy needs agent-based modelling''~\cite{farmer2009economy}.

However, despite the excitement in academic circles, adoption of ACE has been slow in real-world contexts, for real-world economic policy design~\cite{Hajinasab2014}. Several ambitious projects have been pursued, such as EURACE~\cite{Deissenberg_2008}---which had an intention to model the \emph{entire} European economy as an agent-based simulation---but these projects have been distinctly limited in scope and applicability, facing a series of challenges~\cite{dawid2011agent, LeBaron2008}.

This review, then, aims to synthesize from the literature four core stumbling blocks faced by ACE techniques, in a hope to understand the reasons for its slow uptake. Thereafter, it revisits Reinforcement Learning (RL) for the discipline, as originally envisaged by Holland and Miller~\cite{holland1991artificial}, where agents are \emph{adaptive}. Accordingly, we assess whether any of the identified challenges of ACE can be addressed by RL, by considering the prominent research from the field. The core question is thus: has the development in modern RL algorithms unlocked a new frontier for computational economic policy design, transcending the historical challenges faced by agent-based modelling?

With this research question in mind, we now highlight a distinct scope for this review. Importantly, this paper is notably cursory, and does not claim to delve into any of the ideas comprehensively. There exists a plethora of research into the intersection of AI and Economics---the reader is encouraged to see the broad introduction by Furman and Seamans~\cite{Furman_2019}, or various other works~\cite{Agrawal2019, Mosavi2020, Zhang2021} for in-depth examples and analyses. In this review, only \emph{agent-based} computational modelling approaches are considered, and only in a handful of economic settings; moreover, the only paradigm of AI considered is that of RL, as a subset of Machine Learning, and this is done only insofar as it is relevant to ACE's challenges. While there are many other fascinating avenues to explore in these domains, they remain outside of this report's scope. 

This review proceeds as follows. Section~\ref{sect:abcm-intro} introduces the field of Agent-Based Computational Economics (ACE) and its key concepts. Four challenges facing ACE are synthesized from the literature and presented. Each challenge is supported by research, either through theoretical analyses or actual ACE implementation reports, and the high-level criticisms are summarised. Section \ref{sect:can-rl-help} then dives in Reinforcement Learning (RL) in the context of computational economics, and presents a `Scorecard'---where each of the four challenges is discussed with the state of contemporary RL in mind. We remark that of the four challenges presented, RL might assist in just one regard: the need for an explicitly-modelled agent. Next, Section \ref{sect:case-studies} presents a short table of case studies in which RL has indeed been attempted for use in computational economics. Finally, conclusions are drawn, and recommendations are made for future research.

\section{Traditional Agent-Based Computational Economics\label{sect:abcm-intro}}

\subsection{Overview}
Agent-based modelling, as its name implies, is centred around so-called ``agents''---entities in a computer model which have prescribed behaviours, codified goals, etc., and can interact with each other in various ways~\cite{farmer2009economy}. An economic model using these agents may, for example, simulate an array of stock brokers, each trying to maximise their returns in a market. Akin to the hypothetical \emph{Homo Economicus}---the perfectly rational ``Economic Man'' invoked by many economists---Parkes and Wellman describe an agent in computer-based simulation as \emph{Machina Economicus}~\cite{parkes2015economic}. 

The popular and prevailing approach to economic modelling is based on so-called `Dynamic Stochastic General Equilibrium' (DSGE) models. Here, the focus is not on the individual agent entity, but an \emph{aggregate} of the entities in a system---a ``top-down'' approach~\cite{tesfatsion2006handbook}. Though useful for its analytical tractability, DSGE models make assumptions of equilibrium and a ``perfect'' environment~\cite{farmer2009economy}, and have thus struggled with representing complexity in real-world systems, such as economic crashes and crises. Agent-based modelling was championed as an antidote to this issue, by introducing \emph{heterogeneity}---a ``bottom-up'' approach where each agent can have unique characteristics. This technique is part of a broader trend towards the notion of ``Complexity Economics''~\cite{arthur2021foundations}, in a hope to model the nuances of the real-world more accurately. It is believed that through this approach, so-called ``emergent'' behaviour can arise~\cite{Epstein2012}, allowing for far richer and more realistic analysis of systems. Gilbert and Hamill~\cite{hamill2015agent} describe this movement as bridging ``the gap between micro and macro''---meaning between microeconomics, focused on the individual and its microcosm, and macroeconomics, focused on system-wide trends and phenomena.

ACE does seem to have a host of advantages and prospects~\cite{van2012agent}, and has been supported by many prominent voices in the field of economic modelling~\cite{tesfatsion2006handbook, farmer2009economy, holland1991artificial, parkes2015economic}. However, the discipline is not without challenges~\cite{macal2016everything}, and adoption of ACE has been slow for real-world projects~\cite{Hajinasab2014}. DSGE, despite its flaws, has been utilized far more regularly in actual policy design and analysis scenarios~\cite{farmer2009economy}. This section tries to understand why this is the case, by identifying the key challenges of ACE, as drawn from the literature.

\subsection{Core Challenges\label{sect:abcm-challenges}}

\subsubsection{Computational Requirements.}

The first challenge identified in the literature is the intense computational requirements of complex agent-based models. Work has been done to mitigate such difficulties via efficient scaling solutions (e.g.~\cite{Parry_2011}); but the essence remains: the existence of many agents creates exponentially-many inter-agent interactions, which is further aggravated when such interactions are themselves complex or varied~\cite{ajelli2010comparing}. The clearest example of this is in one of the most ambitious agent-based economic simulations to-date: the EURACE model~\cite{Deissenberg_2008}. Conceptually, researchers aimed to model the \emph{entire} European economy, incorporating an array of heterogeneous actors, including households, firms, banks, etc., each with unique behaviours and objectives.

Clearly, the project was enormous, and required effort from many academics over a sustained period. Nominal success was achieved, in narrow scenarios: for example, Dawid et al.~\cite{dawid2008skills} used the model to study the impacts of various economic policies in the labour market. Alas, several years later, Dawid himself admitted that the sheer scale of the task introduces new problems~\cite{dawid2011agent}, which were not fully anticipated. Even when embracing a massively-parallelised approach, the system has to contend with an appropriate segmentation of the problem, communication between parallel threads, and so on---all the standard challenges of concurrent code. These thoughts are shared by LeBaron and Winker~\cite{LeBaron2008}, saying that agent-based modelling, as attractive as it may seem, is ``time consuming.''

\subsubsection{Agent Representation.}

The next identified challenge is a commonly cited one: how should we represent our agents in the model? Van Dam et al.~\cite{van2012agent} call this the \emph{Model Narrative}: ``which agent does what, with whom, and when.'' In the literature, this is also referred to as the \emph{Discrete Choice Model} (DCM) for the agents~\cite{Hajinasab2014}. Holm et al.~\cite{Holm2016} laments that the DCM formulation is often uninformed and ad~hoc. Furthermore, Hajinasab et al.~\cite{Hajinasab2014} regard the oversimplification in DCM as the primary issue for inaccuracies in the resulting ACE models. Holland and Miller~\cite{holland1991artificial}, too, were aware of this issue when first presenting their idea of Artificial Adaptive Agents. In fact, they view it as an inherent challenge of modelling agents with bounded rationality, noting that ``there is only one way to be fully rational, but there are many ways to be less rational.'' Haldane and Turrell~\cite{Haldane_2018} see this challenge as well, questioning which behaviours ought to be incorporated at the agent-level, and which can be safely omitted. Because of these pre-defined and explicit behavioural rules, D'Orazio~\cite{DOrazio2017} further argues that, by hand-coding the agent's nature, the researcher potentially incorporates significant bias into the simulations, thus rendering them less meaningful or useful.

\subsubsection{Interpretability: Verification, Validation, and Reporting of Results.}

The third identified challenge is one which is evident across the computer-based modelling landscape: interpretability---how do we make sense of the outcomes of our models? When looking at agent-based modelling, there seem to be three aspects of interpretability to consider. Firstly, verification: how can we ensure the implemented model corresponds to the conceptual model designed beforehand~\cite{galan2009errors}? That is, how can the programmers ensure their code is bug-free, with respect to their initial brief? Secondly, validation: how can we ensure the model actually reflects the underlying reality it is trying to represent~\cite{carson2002model, Fagiolo_2007}? Finally, how do we report our results in a meaningful way? Srbljinovi\'c and Skunca call this ``explanatory opacity''~\cite{srbljinovic2003introduction}, and note that it is especially important in the current discussion: many of the economists using a given model would likely be unaware of how it works or represents the world, in code. Indeed, Haldane and Turrell~\cite{Haldane_2018} state that ``modellers must also ask themselves how the results of complex simulations of complex systems can best be communicated to researchers and policy-makers alike.''

In this vein, Westerhoff and Franke~\cite{westerhoff2012agent} note how we are certainly not at a stage where the model can operate without human intervention. Despite being very optimistic about the possibilities of ACE, they note the importance of human ratification for sensible models.

\subsubsection{Model Stochasticity.}

The final challenge identified across the literature is sometimes seen as a benefit, rather than a drawback. That being, the stochastic nature of model outputs. For a simple example, consider a result provided by Raberto et al.~\cite{Raberto2008}, in which three unique trajectories of interest rates are proposed by the model, despite starting with \emph{identical} initial conditions. Haldane and Turrell~\cite{Haldane_2018} accordingly denote agent-based models as machines for ``generating many alternate realizations of the world.'' In some senses, this is desirable: researchers can run the simulations over and over, and each time envisage a novel reality, each with unique implications and insights.

Not all authors see this property positively, though. LeBaron and Winker~\cite{LeBaron2008}, for instance, point out that significantly different policy recommendations could be concluded from multiple iterations of a simulation, despite little to no changes in the underlying model. They then question the reliability of such simulations as a policy-design tool.

\section{Can Reinforcement Learning Help?\label{sect:can-rl-help}}

\subsection{Overview}
The science of Reinforcement Learning (RL) has been extant for many decades, but the field has recently seen an explosion of advances. Largely, this transformation has been spurred on by the integration of Deep Learning into the domain~\cite{Silver2016}, denoted as \emph{Deep Reinforcement Learning} (DRL). This review does not delve deeply into the theoretical aspects of RL, and simply accepts that significant progress has been made in the field---the reader is encouraged to see the canonical text by Sutton and Barto~\cite{SuttonBarto2018} for a thorough discussion of the discipline, as well as the survey by Arulkumaran et al.~\cite{Arulkumaran2017} for recent advances. This section, instead, simply contemplates whether the domain of RL can help overcome any of the aforementioned challenges of agent-based modelling. Note that there is a close link between the two approaches, where the former also defines an \emph{agent-based} scenario, but importantly with agents that can \emph{learn} over time. We are asking, then, does this \emph{adaptive} nature of RL help ACE?

Hitherto, it seems that there are no explicit studies in the literature comparing the challenges of agent-based techniques to the features of RL, and certainly not in the context of economic policy design. As a result, there lacks a clear systematic approach comparing the disciplines, which makes definitive conclusions difficult. Instead, we aim to amalgamate conclusions from various disparate RL use-cases, unrelated to economics or agent-based modelling, and then infer the possible traits of RL in such settings. This is not to say RL will \emph{necessarily} solve the issues mentioned; instead, it postulates the aspects of agent-based modelling in which RL may assist, thus motivating for these aspects as domains for future research.

\subsection{Comparative Scorecard}
This brief `scorecard' considers each of the four challenges of agent-based modelling outlined previously, and suggests whether RL has the potential to overcome them.

\subsubsection{Computational Requirements.}
The first challenge with ACE cited from the literature was the sheer computational challenge of modelling many individual agents in simulation. Indeed, it was recognised as a key hurdle for real-world projects, such as EURACE~\cite{Deissenberg_2008}. Could the introduction of \emph{learning}---such that agents are adaptive---reduce this computational burden?

Quite clearly, no. As pointed out recently by Ceron and Castro~\cite{pmlr-v139-ceron21a}, researchers in RL are, too, grappling with the computational difficulties of training large and complex models. In their surveillance of the literature for DRL in Autonomous Driving, Kiran et al.~\cite{Kiran2021} support this notion, citing how expensive many DRL algorithms can be. One can deduce that introducing \emph{adaption} to an agent in an economic model, moving from the traditional agent-based approach to an RL approach, can only \emph{increase} the computational requirements of the system~\cite{Epstein2012}, since the associated complexity must grow monotonically. Moreover, based on the extant RL literature, it seems that this additional cost may even be significant. Accordingly, we strongly doubt whether RL can assist ACE in this regard.

\subsubsection{Agent Representation.}
The next challenge identified by many authors is that of agent representation. It is here that RL has shown the most promising results. Consider first the great success achieved by Silver et al.~\cite{Silver2016} in the creation of `AlphaGo'---an algorithm that, for the first time, defeated a professional human player in the ancient game of Go. Wang et al.~\cite{wang2016alphago} reflect on the technological implications of this advancement, seeing it as a watershed moment for a new era of computational intelligence in complex systems. It is worth noting, however, though this algorithm \emph{did} rely on RL, there was still a significant degree of agent design from humans, based on empirical game data. Arguably, then, in the context of in ACE, the development of AlphaGo does not solve the fundamental problem of improving agent representation.

The developers of AlphaGo shared this sentiment, insisting that we ought to create algorithms that can learn \emph{tabula rasa}---that is, in the absence of any preconceived notions, ideas, goals. And indeed, just one year later, Silver et al. created AlphaZero~\cite{Silver2017}---which mastered the same game of Go, but \emph{solely} using RL, without any human guidance or domain knowledge. Not long after this, an even bigger step was taken in the creation of MuZero~\cite{Schrittwieser2020}. Here, Schrittwieser et al. developed an agent that could master several games, including Go, \emph{without} being told the underlying dynamics of the system in question. That is, the agent was not told how it should play the game, nor how the game even works---it had to learn such dynamics from experience.

The motivation for this pursuit is that ``in real-world problems, the dynamics governing the environment are often complex and unknown''~\cite{Schrittwieser2020}---notice that this is the precise problem faced by agent-based modelling in economic contexts. Despite not knowing the system rules \emph{a priori}, the MuZero algorithm was able to achieve state-of-the-art performance in playing the games, unlocking a new conceptual framework for how we can train computer agents. Silver et al.~\cite{Silver2021} discuss the philosophical dimensions of this training approach, in a treatise entitled ``Reward is enough.'' Whether the maximisation of reward is, in fact, solely enough for the realisation of intelligence, or something more nuanced is required (for example, as suggested by Juechems and Summerfield~\cite{Juechems_2019}), the point remains: advances in RL have revealed new possibilities for the creation of intelligent agents---ones which have not been hard-coded by humans.

\subsubsection{Interpretability: Verification, Validation, and Reporting of Results.}
Interpretability is another major issue in the context of agent-based modelling. By enabling agents the ability to learn, can we improve the so-called ``explanatory opacity''~\cite{srbljinovic2003introduction} of the model? Unfortunately, it seems not. Dulac-Arnold~\cite{DulacArnold2019} see this characteristic as one of the core issues of real-world RL. They mention that those relying on RL algorithms desire ``explainable policies and actions,'' which is something that has hitherto been difficult to provide. In the context of RL in neuroscience, Eckstein et al.~\cite{Eckstein2021} mention the importance of attaining a nuanced understanding of RL models and their outcomes, for the sake of moving the field forwards towards being a practical tool, used in real situations.

Though progress is being made in this regard~\cite{DulacArnold2019}, it is clear that RL faces the same issue of interpretability as agent-based modelling in the economic context. Indeed, it is clear that simply allowing agents in a model to learn does not unlock new possibilities of interpreting any complex and multi-faceted dynamics that emerge.

\subsubsection{Model Stochasticity.}
Finally, regarding the stochasticity of agent-based models: does the introduction of \emph{adaption} in the agents' behaviour improve the degree of determinism in our results? Perhaps obviously: no. In fact, many RL algorithms and methodologies introduce \emph{new} aspects of stochasticity~\cite{SuttonBarto2018}. In a landmark paper, Henderson et al.~\cite{Henderson_Islam_Bachman_Pineau_Precup_Meger_2018} lament that the Deep RL landscape, despite experiencing exciting progress, is facing a serious challenge: being able to reproduce results reliably. Their motivation is largely framed around reducing the duplication of work: reproducible results are easier to build off in academic research, than starting from scratch every time. Nonetheless, their argument holds the same kernel of truth as was the case for ACE: multiple iterations of an RL model will almost certainly produce unique results. Once again, this may be viewed positively~\cite{Haldane_2018}, but it certainly does not solve the issue faced by ACE.

\subsection{Remarks}
This section considered the identified problems evident in Agent-Based modelling techniques, and contemplated whether advances in RL have the possibility to assist. Of the four broad categories discussed, it was found that only one of them could conceivably be solved---or at least improved---via the adoption of RL. That being, the explicit representation of agents in the system. The other issues, the literature suggested, were either not solved or were even aggravated when using RL.

\section{Case Studies\label{sect:case-studies}}
We provide in Table~\ref{tab:case-study} a short summary of four case studies where RL is used in agent-based simulations, broadly for topics in economic modelling. Through these examples, the core takeaway of the review---that RL can assist in the regard of agent-based representations---is affirmed. 
\input{cs_table.tex}

\section{Conclusions}
This review aimed to answer the question of whether advances in Reinforcement Learning (RL) could overcome the difficulties historically faced by agent-based methods, in the context of computational economic modelling. It began by discussing agent-based computational economics (ACE) as a discipline, and synthesized from the literature four unique challenges of the field. It is reasoned that these four challenges are, at least partly, why ACE has not been adopted in mainstream economic modelling toolboxes in any significant way---despite high hopes from its proponents~\cite{holland1991artificial}. With these four challenges in mind, then, the viability of RL was considered as a potential solution for each. Essentially, this was a consideration whether the ability for the agents to \emph{learn} in the model mitigated the challenges whatsoever.

Drawing from theoretical and practical analyses alike, it was found that just one of the challenges could potentially be overcome using RL: the problem of ``agent representation''---that is, the way the agent is modelled in the system. The other challenges, such as computational complexity, were either unaffected, or even aggravated, via the use of adaptive, RL-based agents. To reinforce these ideas, four case studies were then given briefly, with examples of RL being successfully used in agent-based models, and indeed, assisting with the complexities of agent representation.

This review was notably cursory, with the aim to provide a straightforward yet wide surveillance of the available literature for the topic in question. Consequently, there is ample scope for further discussion. For starters, a more-thorough analysis of RL methods in ACE deserves attention, especially one that methodically compares each approach directly to the other---including the associated costs, accuracy achieved, etc. Moreover, there are far more aspects of computational economics and finance worth exploring in this context, as well as many alternative Machine Learning paradigms and strategies. Clearly, then, there exists a multitude of combinations between these two disciplines, all of which should at least be considered. Finally, more ambitiously, larger projects---in the spirit of EURACE~\cite{Deissenberg_2008}---should be investigated, now leveraging the benefits of modern RL techniques. The potential for such research is rich and plentiful.

\subsubsection{Acknowledgements.} 
Thanks to Dr Pavlos Andreadis for useful discussions at the start of this enquiry.

\bibliographystyle{splncs04}
\bibliography{main}       %
\end{document}

%% file: cs_table.tex
\begin{table}[htbp]
\centering
\caption{Case Studies of RL in Economic Modelling}
\label{tab:case-study}
\resizebox{\textwidth}{!}{%
\begin{tabular}{|l|l|}
\hline
\textbf{Modelling...} & \textbf{Examples} \\ \hline
Dynamic Pricing in Markets & \begin{tabular}[c]{@{}l@{}}Kutschinski et al.~\cite{kutschinski2003learning}: DMarks II platform\\ Radhakrishnan et al.~\cite{Radhakrishnan2015}: Electricity market\\ Ghasemi et al.~\cite{Ghasemi2020}: `Prosumer' microgrid electricity model\end{tabular} \\ \hline
Adaptive Incentive Scenarios & \begin{tabular}[c]{@{}l@{}}Yang et al.~\cite{yang2022}: Optimal tax policy\\ Zheng at al.~\cite{zheng2020ai}: Optimal tax policy (The `AI Economist')\\ Castro et al.~\cite{castro2020estimating}: Liquidity allocation in high-value payment systems\end{tabular} \\ \hline
Fiscal Policy & \begin{tabular}[c]{@{}l@{}}Hinterlang and Tänzer~\cite{hinterlang2021optimal}: Optimal interest rates\\ Chen et al.~\cite{chen2021deep}: Monetary model with household agents\end{tabular} \\ \hline
Sequential Social Dilemmas & \begin{tabular}[c]{@{}l@{}}Leibo et al.~\cite{leibo2017multi}: Temporally-extended dilemmas\\ Gléau et al.~\cite{gleau2022tackling}: Asymmetric and circular dilemmas\\ Eccles et al.~\cite{eccles2019learning}: Reciprocal behaviour\end{tabular} \\ \hline
\end{tabular}%
}
\end{table}